# Runtime Burden Allocation for Structured LLM Routing in Agentic Expert Systems: A Full-Factorial Cross-Backend Methodology

Zhou Hanlin[1,2]; Chan Huah Yong[1]*

[1] School of Computer Sciences, Universiti Sains Malaysia (USM), Penang, Malaysia
[2] Xiamen Software Vocational & Technical College, Xiamen, China
* Corresponding author: hychan@usm.my

**Abstract**

Structured LLM routing is often treated as a prompt-engineering problem. We argue that it is, more fundamentally, a systems-level burden-allocation problem.

As large language models (LLMs) become core control components in agentic AI systems, reliable structured routing must balance correctness, latency, and implementation cost under real deployment constraints. We show that this balance is shaped not only by prompts or schemas, but also by how structural work is allocated across the generation stack: whether output structure is emitted directly by the model, compressed during transport, or reconstructed locally after generation.

We evaluate this formulation through a comprehensive full-factorial benchmark covering 48 deployment configurations and 15,552 requests across OpenAI, Gemini, and Llama backends. Our central finding is consequential: there is no universal best routing mode. Instead, backend-specific interaction effects dominate performance. Modes that remain highly reliable on Gemini and OpenAI can suffer substantial correctness degradation on Llama, while efficiency gains from compressed realization are strongly backend-dependent.

Rather than presenting another isolated model comparison, this work contributes a deployable framework for reasoning about structured routing under heterogeneous backend conditions. We provide a cross-backend evaluation methodology and practical deployment guidance for navigating the correctness–cost–latency frontier in production-grade agentic expert systems.

*Keywords: expert systems; structured routing; agentic systems; backend compatibility; function calling; local reconstruction*

## 1. Introduction

Large language models are increasingly used as control components inside compound AI systems rather than as stand-alone text generators (Romain & Sarath, 2024). In many real deployments, the operational event that matters most is not the wording of a reply but the control record that determines where the request should go next, whether memory should be queried, whether a tool should be called, and whether a developer-oriented subsystem should be activated (Cancedda et al., 2023; Li et al., 2023; Wu, Wang, & Zhang, 2023; Yao et al., 2022; Zhou & Chan, 2026). This front-door controller role is especially important in agentic expert systems, where the quality of routing can determine whether the downstream software stack behaves correctly, whether expensive tool calls are wasted, and whether interactive latency remains within application-level service targets (Kernan Freire et al., 2024; Romero, Zimmerman, Steinfeld, & Tomasic, 2023).

For such systems, correctness is inherently structured. A Director-like controller must produce a record that is machine-readable, semantically correct, stable under interrupted or stateful dialogue, and operationally fast enough for the host workflow. If the route label is wrong, the wrong subsystem may answer. If the record is malformed, the orchestration layer may fail (Greshake et al., 2023; Tang et al., 2024). If the response arrives only after the downstream queue has already stalled, the user will experience the router as slow even if the underlying model is nominally capable. These requirements make structured routing different from free-form question answering. The problem is not merely "can the model understand the task?" but also "can the runtime package that understanding into a reliable control artifact?".

Recent literature already shows that these concerns are not peripheral. Work on function calling, tool use, and tool-learning benchmarks demonstrates that model reasoning and system integration interact in non-trivial ways (Qin et al., 2024). Research on structured output generation similarly shows that schema compliance is not free: strong models remain sensitive to formatting burden, schema complexity, constrained generation, and post-processing decisions (Tam et al., 2024). Routing and cascade studies add another perspective by showing that multi-model systems have cost-quality trade-offs that are not visible from isolated model evaluation alone (Sienicki, 2024; Yue, 2024). Together, these studies imply that system behavior depends on much more than "the model" in the abstract.

What remains underdeveloped is a more specific question that matters greatly in practice: if the routing task is fixed, how much of the observed behavior is determined by runtime packaging rather than by model identity alone? Real system builders regularly face design choices such as whether to stream or not stream, whether to require final JSON from the model or accept a shorter intermediate code, whether to relax generation budget, and whether to push part of structural realization into deterministic local software. These decisions are sometimes treated as low-level engineering details. However, if they materially alter route correctness, structure validity, and latency, they should be elevated to first-class expert-systems design variables.

This manuscript addresses that gap by reframing structured LLM routing as a runtime burden-allocation problem. The central idea is that every deployment profile allocates structural work differently across the model and the surrounding software. Some profiles ask the model to produce the final structured record. Others compress the emitted representation and reconstruct the final schema locally. Some rely on streaming transport, others do not. These choices alter the "burden" placed on the model and therefore alter the observed control behavior.

The contribution is intentionally methodological rather than purely algorithmic. The paper does not claim a new foundation model or a new routing learner. Instead, it introduces a compact theoretical abstraction and tests it against a controlled full-factorial benchmark consisting of 48 deployment combinations, 324 requests per combination, and 15,552 routed requests overall. The benchmark spans four runtime modes, three backend families, two constraint settings, and two transport conditions. The resulting evidence permits a stronger claim than an ordinary benchmark comparison: there is no backend-independent optimum for structured routing runtime packaging. Instead, backend–mode compatibility emerges as a first-order systems effect.

The rest of the paper proceeds as follows. Section 2 reviews related work on agentic control, function calling, structured output generation, and routing evaluation. Section 3 develops the runtime burden-allocation framework and states the methodological propositions that guide the study. Section 4 describes the full-factorial benchmark, task design, metrics, and analysis protocol. Section 5 presents the empirical results, including backend × mode interactions, efficiency–correctness trade-offs, and failure taxonomy. Section 6 discusses the methodological implications, deployment rules, external-validity boundaries, and remaining limitations. Section 7 concludes.

## 2. Related work

This section positions the study against four adjacent literatures: LLMs as control components, tool-use reliability, structured output generation, and routing or cascade evaluation. The goal is not to review the entire LLM systems field, but to isolate the strands that matter most for compact structured routing in expert systems.

### 2.1 LLMs as control components in expert systems

A growing body of work treats LLM outputs not simply as answers but as intermediate control artifacts. Toolformer, ReAct, Gorilla, HuggingGPT, AutoGen-style orchestration, and related systems demonstrate that language models can select actions, parameterize external APIs, and coordinate external software rather than only produce narrative text. Broader prompting, predictability, and self-improvement work also shows that intermediate reasoning structure and revision strategy can materially change downstream behavior (Alon et al., 2023; Bosma et al., 2022; Chen et al., 2024). In expert systems this control view is especially relevant because the "answer" may be only one step in a wider workflow. The upstream router, planner, or controller often determines whether the rest of the pipeline even has a chance to succeed.

This perspective aligns well with the aims of *Expert Systems with Applications*, which emphasizes intelligent systems not only as algorithms but as deployable systems with design, testing, implementation, and management implications (Ling et al., 2025). A Director router fits that orientation directly. It is an expert-systems component whose purpose is to transform natural-language requests into structured operational choices. What matters therefore is not just model intelligence in the abstract, but the systems reliability of the emitted control record.

### 2.2 Function calling and tool-use reliability

A second stream examines LLM function calling and tool use more directly (Daiber, Maricato, Sinha, & Rabinovich, 2025). Recent studies show that tool-use evaluation has moved beyond the question of whether models can invoke tools at all. The current focus is whether they can do so reliably, under multi-turn settings, with realistic ambiguity, and with proper argument structure (Chen et al., 2024). Recent benchmarks further stress multilinguality, tool retrieval, open-world tool selection, and robustness under perturbation rather than narrow single-turn success/failure framing (Erdogan et al., 2024; Saha et al., 2024). This shift is directly relevant to runtime packaging because a controller may fail even when the underlying semantic intent is partly correct.

This literature is directly relevant because structured routing is often the precursor to tool calling. Before a system decides *which tool* to invoke, it frequently needs to decide *which subsystem or branch* should handle the request at all. The router therefore occupies a position one layer earlier than tool execution. If the route is wrong, downstream tool execution quality becomes moot. Yet tool-use studies usually emphasize end-to-end outcomes or argument accuracy rather than the reliability of a compact routing record itself. The present study complements that literature by isolating the routing layer.

### 2.3 Structured output generation

A third stream studies structured output generation in its own right. This is the closest literature to the present problem because Director routing requires a small schema whose fields must be correctly emitted, parsed, and acted upon. Recent benchmark efforts show that structured generation quality remains uneven across formats and constraints. SLOT proposes a model-agnostic transformation approach that repairs unconstrained outputs into structured forms (Shen et al., 2025), directly illustrating that one way to improve practical reliability is to move some structural work out of the generation loop. Schema reinforcement learning (Ayala & Bechard, 2024) further suggests that model-side structured generation can be improved, but only through explicit optimization of schema-conditioned behavior (Tavanaei et al., 2024). Work on reasoning over structured environments and structured memory likewise indicates that schema-bearing intermediate states can help only when representation fidelity is preserved (Cheng et al., 2024).

These studies collectively imply that structure itself imposes cost. When a model is asked to infer the correct decision *and* serialize that decision into the final target schema, some portion of the error may come from the structure realization burden rather than the underlying semantic choice. That insight is foundational for the present paper.

### 2.4 Routing, cascading, and systems-level choice

A fourth stream comes from routing and cascading research. Multi-LLM routing studies show that model selection can be framed as a cost-quality allocation problem , while benchmarks such as RouterBench and RouterEval attempt to standardize the evaluation of such routers. Routing-and-cascading work further shows that cost-quality frontiers depend strongly on router design and uncertainty estimation rather than on a single dominant model. Related agent benchmarks such as ASSISTANTBENCH, SWE-Bench, LegalAgentBench, and Hephaestus reinforce the same lesson: orchestration quality and tool-grounded control must be evaluated separately from raw language fluency (Jimenez et al., 2023; Yoran et al., 2024). These studies are important because they demonstrate that routing deserves analysis in its own right rather than being treated as a trivial wrapper around model inference. However, most of this work varies the model pool or routing learner. Much less attention is paid to runtime packaging of the control record itself when the task remains fixed(Liu et al., 2023).

The present study occupies that gap. Rather than asking which model should answer a query, it asks how the same structured control task changes when runtime packaging changes. That distinction matters because many practical teams do not want to maintain a large heterogeneous model pool. They instead need to understand how much performance can be gained or lost by changing the runtime interface around a fixed backend family.

### 2.5 Research gap

The literature therefore reveals three gaps. First, structured routing has clear expert-systems relevance but remains under-analyzed as a systems component. Second, structured output work shows that schemas matter, but rarely frames the issue as burden allocation between model-side generation and deterministic local software. Third, routing studies emphasize model selection more than runtime packaging. This paper addresses all three by proposing a burden-allocation framework for structured routing and grounding it in a full-factorial cross-backend benchmark.

## 3. Runtime burden-allocation framework

### 3.1 Structured control formulation

The central premise of this paper is that Director routing should be modeled as structured control rather than as abbreviated conversation. Let the incoming request be x and the desired control record be y. In the benchmark studied here, y contains five fields: a route label, a confidence value, a memory flag, a tool flag, and a short reason string. The route set is fixed and compact: R = {chat, task, dev, doc}. A deployment is useful only if the emitted record is both operationally valid and decision-correct.

To represent the observable behavior of a deployment, define the observed outcome vector

$$O = (FC, RA, SR, LAT, TOK)$$

where FC denotes format compliance, RA denotes routing accuracy, SR denotes state retention under applicable interrupted/stateful cases, LAT denotes response latency, and TOK denotes token consumption. For a backend b and runtime configuration m, the observed behavior can be expressed as

$$O_{\{b,m\}} = \Phi(b, \psi_m, D)$$

where D is the task distribution and $\psi_m$ is the runtime burden-allocation profile of mode m. This equation is not intended as a closed-form predictor. Its purpose is methodological: it makes explicit that the observed result is a joint function of backend family, runtime burden profile, and task distribution. The relevant implication is that structured routing quality cannot be attributed to model identity alone.

### 3.2 Burden-allocation dimensions

The runtime burden-allocation profile $\psi_m$ is defined by three dimensions.

(1) Serialization burden.
This dimension captures how much schema construction the model must perform inside its generation loop. High serialization burden means the model must emit the final machine-readable structure, with field names, delimiters, typing cues, and appropriate syntax. Low serialization burden means the model can emit a smaller intermediate code from which deterministic software reconstructs the final schema (Khattab et al., 2023).

(2) Transport semantics.
This dimension captures whether the control record is delivered via streaming or non-streaming transport. Streaming can improve time-to-first-token or subjective responsiveness in conversational settings, but in control settings the partial output may have no operational value until the final record is complete.

(3) Locus of structure realization.
This dimension distinguishes whether the final machine-readable artifact is produced directly by the model or reconstructed by local deterministic software. This is conceptually distinct from serialization burden. A runtime may use a compact intermediate format but still require substantial semantic precision from the model, or it may preserve a rich target schema but use strong post-processing to stabilize it.

Together, these dimensions define a burden-allocation profile. The four runtime modes in the benchmark are simply concrete instantiations of these dimensions.

### 3.3 Compatibility principle

From these dimensions follows the compatibility principle:

> A runtime profile should be judged relative to a backend-conditioned utility frontier, not by a backend-independent notion of optimality (Wallace et al., 2024).

The principle means that moving structural work out of the model loop can be beneficial, but the benefit depends on whether the backend remains semantically stable under the compressed representation (Opsahl-Ong et al., 2024). If the compressed intermediate code becomes ambiguous or fragile for a given backend, local reconstruction may improve cost while harming correctness. Conversely, if the backend handles compact codes reliably, moving recoverable syntax out of the generation loop can reduce burden and improve efficiency.

This principle is what converts the paper from a simple benchmark into a methodology. It predicts that the same runtime packaging choice may have different effects across backend families.

### 3.4 Methodological propositions

To make the framework testable, the paper advances three propositions.

Proposition 1 (Interaction proposition).
For structured routing tasks, backend × runtime-mode interaction will be a first-order determinant of observed control quality.
This proposition is supported if interaction terms are statistically significant and substantial for correctness-oriented metrics such as RA, SR, and FC.

Proposition 2 (Recoverable-structure proposition).
When the target schema is compact and enumerable, moving recoverable structure realization from the model to deterministic local software can improve efficiency, but only if the compressed intermediate representation remains semantically stable for the chosen backend.
This proposition predicts that compressed local reconstruction will often improve latency and token usage, but that correctness gains will be backend-conditioned rather than universal.

Proposition 3 (Actionable-latency proposition).
For structured control tasks, full-response latency is a more decision-relevant measure than time-to-first-token, because no partial output is actionable until the record is complete.
This proposition predicts that streaming will have limited value for control quality and may contribute only a secondary effect on operational latency.

### 3.5 Practical interpretation

The framework intentionally stops short of claiming a universal predictive theory. It is a lightweight operational theory designed to organize design choices, expose interaction effects, and guide deployment decisions. Its value lies in testability and transferability: it provides a vocabulary and experimental protocol for comparing runtime packages in any compact structured routing problem , even if the precise optimal profile changes from system to system.

### 3.6 Applicability domain and falsifiability

The framework is intended to be falsifiable rather than merely descriptive. It applies most directly when four conditions hold: (i) the route ontology is compact and enumerable; (ii) the emitted record is a control artifact rather than a final user answer; (iii) at least part of the target schema is recoverable by deterministic local software; and (iv) downstream execution cannot proceed until the full control record is available. Within that domain, the theory predicts large backend x mode interaction, efficiency gains from offloading recoverable structure, and limited operational benefit from streaming.

The framework would be weakened by contrary evidence of three kinds. First, if backend x mode interaction were consistently negligible once backend main effects were controlled, burden allocation would not deserve first-order status. Second, if compressed local reconstruction preserved correctness almost uniformly across backend families, compatibility would be unnecessary as a design concept. Third, if time-to-first-token or transport semantics dominated full-response latency in compact control tasks, the actionable-latency proposition would fail. These explicit failure conditions make the abstraction methodologically stronger than an unconstrained narrative explanation.

Figure 1 summarizes the burden-allocation framework that motivates the benchmark design and the interpretation of runtime-package behavior.

**Compatibility principle: no backend-independent optimum**

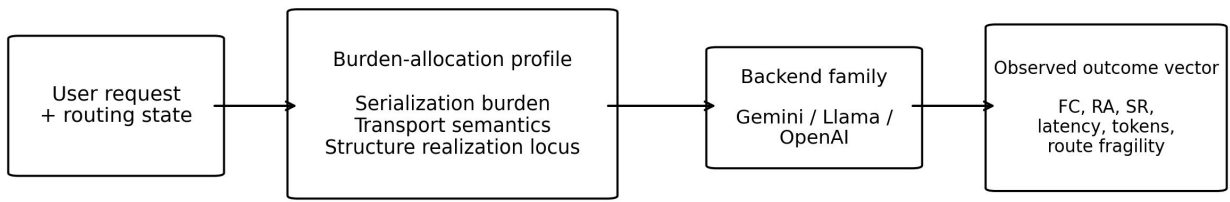

*Figure 1. Runtime burden-allocation framework and the three dimensions used to interpret the benchmarked deployment packages.*

## 4. Experimental design

### 4.1 Full-factorial benchmark matrix

The empirical study evaluates 48 deployment combinations defined by four runtime modes, three backend families, two constraint settings, and two transport settings. The factorization is therefore:

- Mode: MJ, SJ, MJS, MCLR
- Backend: Gemini, Llama, OpenAI
- Constraint: Limited, Unlimited
- Transport: non-stream, stream

Each combination is evaluated on 324 requests, yielding 15,552 requests overall. The archived benchmark script explicitly encodes the matrix as $4 \times 3 \times 2 \times 2 = 48$ combinations, and the benchmark summary report confirms that all 48 combinations were completed with the same total request count.

## 4.2 Runtime modes

The four modes instantiate different burden-allocation profiles.

MJ is the minimal JSON baseline. It uses a small token budget and requires the model to emit the structured record directly as JSON.
SJ retains direct JSON emission but relaxes the output budget substantially.
MJS preserves the JSON target while adding streaming transport.
MCLR is the compressed-output mode with local reconstruction. It uses a small token budget, plain-text style compact emission, and deterministic local reconstruction to realize the downstream JSON structure.

The critical point is that the modes are not merely arbitrary "prompt variants." They correspond to systematic differences along the burden-allocation dimensions defined in Section 3.

To remove a possible ambiguity, the mode label in this manuscript denotes a historically named burden-allocation package rather than a pure transport state. The factorial transport variable is modeled separately as the execution-time API delivery mode. In other words, mode captures the package-level choice about serialization burden and structure realization, whereas transport overlays stream versus non-stream execution on top of that package.

This distinction matters especially for MJS. The label is retained for continuity with the benchmark implementation, but it should not be read as if the letter "S" exhausts the transport manipulation. In the factorial design, transport remains an orthogonal factor applied at execution time to all mode packages. Table 1 therefore makes explicit which part of the design belongs to the named package and which part belongs to the transport overlay.

Table 1. Runtime-package interpretation and transport orthogonality.

| Mode | Serialization | Realization locus | Historical label | Benchmark transport |
| --- | --- | --- | --- | --- |
| MJ | Low JSON | Model emits final JSON | Baseline name | Stream/non-stream overlay |
| SJ | High-budget JSON | Model emits final JSON | Baseline name | Stream/non-stream overlay |
| MJS | JSON package | Model emits final JSON | Historically stream-associated name | Stream/non-stream overlay |
| MCLR | Compact code | Local deterministic reconstruction | Compression / reconstruction | Stream/non-stream overlay |

## 4.3 Task structure

The routing task is intentionally compact but operationally meaningful. Each request must be assigned to one of four route labels—chat, task, dev, or doc—and accompanied by the auxiliary flags required by downstream orchestration. The prompt pool is stratified to include simple cases, complex or interrupt-sensitive cases, and edge cases. This design is deliberate: a deployment-oriented benchmark should not rely on a single homogeneous test set because different runtime packages may fail under different regimes.

The prompt pool is nevertheless bounded. This is a strength for controlled comparison but also a limitation for external validity. The benchmark should therefore be interpreted as a controlled structured-control study rather than as a universal routing ontology benchmark.

### 4.4 Metrics

The study uses five primary metrics.

- Format compliance (FC): whether the output is parseable and schema-valid.
- Routing accuracy (RA): whether the route label is correct.
- State retention (SR): whether state-sensitive or interrupted cases are handled correctly where applicable.
- Latency (LAT): full-response latency, summarized by p50 and p95.
- Tokens (TOK): total token consumption.

These metrics correspond directly to the observed outcome vector O in the burden-allocation framework (Guo, Chen, & Wang, 2024). They also map naturally to deployment value. FC captures machine-readability, RA captures decision quality, SR captures stability under multi-turn or interruption-sensitive settings, LAT captures operational responsiveness, and TOK captures inference cost.

For workflow interpretation, the paper also reports a derived workflow lower-bound completion (WLC) proxy. WLC is defined using the Fréchet lower bound on the joint event that a request is format-valid, correctly routed, and state-safe: WLC = max(0, FC + RA + SR - 200) for three-event slices, or max(0, FC + RA - 100) when state retention is not applicable. WLC is not a direct downstream task-success metric. It is a conservative operational proxy for the minimum rate at which a downstream expert-system workflow can receive a usable, correctly dispatched, and state-safe control record from the router.

### 4.5 Statistical analysis protocol

The analysis proceeds at the combo level. For each metric, the paper first reports descriptive summaries and backend × mode means. It then applies factorial linear models of the form metric ~ backend * mode + constraint + transport to test the main effects and the backend × mode interaction. Type-II ANOVA is used to summarize F statistics, p-values, and partial eta squared effect sizes for each term. This statistical layer is important because it converts a descriptive comparison into a test of the methodological propositions from Section 3.

Because the benchmark is a controlled systems experiment rather than a sample survey from a naturally occurring population, the inferential goal is not population generalization in the classical sense. Instead, the statistical analysis quantifies how strongly the designed factors explain variation across the 48 deployment combinations. That use of statistics is appropriate for expert-systems design studies: it strengthens causal interpretability of the experimental matrix without overstating domain generalization.

### 4.6 Robustness checks and inferential discipline

Because the experiment is intentionally factorial and deployment-oriented, the primary inferential object is the designed combination rather than an unconstrained stream of naturally sampled user traffic. This affects how the statistics should be interpreted. The ANOVA layer therefore serves two purposes. First, it tests whether the major design factors explain systematic variation in the benchmark outcomes. Second, it quantifies whether the interaction terms are large enough to justify a methodology built around compatibility rather than global ranking.

To avoid overclaiming from purely visual comparisons, the analysis also inspects stability within each backend × mode cell across the four repeated settings generated by constraint and transport. For correctness metrics, many cells are nearly invariant or exactly invariant across the four subconditions, which is itself informative: it shows that the dominant instability in this benchmark is not random within-cell noise, but cross-cell movement induced by the major design factors (He et al., 2025). For latency, where some within-cell variability is expected, the interpretation remains qualitative unless a factor is supported by the factorial model and accompanied by a non-trivial effect size.

Effect sizes are interpreted more strongly than p-values in the discussion. This is important because the 48 combinations are designed, not passively observed. A tiny p-value by itself does not justify a systems claim; a large partial eta squared does. In this study, backend × mode interaction is treated as a first-order effect only when it is both statistically significant and substantively large relative to the other terms. That criterion is met for RA, SR, and FC, and met more moderately for p50 latency. By contrast, transport is not treated as a meaningful design lever for compact control simply because its p-value is weak; it is judged secondary because both the p-values and the effect sizes are small.

A final robustness principle is that route-level slices are inspected alongside aggregate accuracy. Aggregate RA can hide the fact that one operationally critical route is fragile while others remain stable. Since expert-system controllers often protect particularly costly or safety-relevant pathways, route-level fragility is treated as part of deployment validity rather than as a purely diagnostic appendix.

## 5. Results

### 5.1 Descriptive overview

The first high-level observation is that the benchmark does not support a universal "best mode" conclusion. Mean results by backend family and runtime mode are shown below.

Two descriptive patterns are immediately visible. First, MJ and SJ preserve correctness most consistently on Gemini and OpenAI. On Gemini, both MJ and SJ achieve 86.11% RA and 75.00% SR; on OpenAI, MJ and SJ remain above 85% RA with 71.88% SR. Second, MCLR provides clear efficiency gains but these do not translate into universal correctness gains. On Gemini, MCLR reduces token cost from roughly 126k to 45.8k and lowers median latency relative to MJ/SJ, but RA drops to 62.96%. On OpenAI, the same efficiency pattern appears, with median latency around 661.52 ms and tokens around 40k, yet RA falls to 58.49%. Llama is the clearest incompatibility case: MCLR is very fast at 174.32 ms p50 and low in tokens, but FC collapses to 53.40% and RA to 22.84%.

The descriptive picture therefore already supports the compatibility principle. Efficiency-oriented compression can be attractive, but its correctness profile depends strongly on backend family.

Table 2. Backend × mode means with 95% cell-level confidence bounds.

| Backend | Mode | FC% | RA% | SR% | p50(ms) | Tokens |
|---|---|---|---|---|---|---|
| Gemini | MJ | 100.00 ± 0.00 | 86.11 ± 0.00 | 75.00 ± 0.00 | 1153.28 ± 104.86 | 126098 ± 0 |
| Gemini | SJ | 100.00 ± 0.00 | 86.11 ± 0.00 | 75.00 ± 0.00 | 1161.81 ± 86.28 | 126098 ± 0 |
| Gemini | MJS | 99.92 ± 0.15 | 61.03 ± 0.15 | 56.08 ± 0.34 | 1190.01 ± 87.55 | 60133 ± 35 |
| Gemini | MCLR | 100.00 ± 0.00 | 62.96 ± 0.00 | 68.75 ± 0.00 | 1014.09 ± 27.88 | 45771 ± 0 |
| OpenAI | MJ | 100.00 ± 0.00 | 85.65 ± 3.67 | 71.88 ± 3.54 | 1043.71 ± 112.64 | 115810 ± 1019 |
| OpenAI | SJ | 100.00 ± 0.00 | 85.88 ± 3.43 | 71.88 ± 3.54 | 1041.49 ± | 115816 ± 996 |

| Backend | Mode | FC% | RA% | SR% | p50(ms) | Tokens |
|---|---|---|---|---|---|---|
| OpenAI | MJS | 100.00 ± 0.00 | 61.11 ± 2.82 | 51.91 ± 7.69 | 1032.27 ± 126.44 | 51950 ± 2561 |
| OpenAI | MCLR | 100.00 ± 0.00 | 58.49 ± 1.67 | 49.83 ± 4.53 | 661.51 ± 36.59 | 40036 ± 6150 |
| Llama | MJ | 97.14 ± 0.15 | 82.33 ± 0.15 | 68.75 ± 0.00 | 231.89 ± 3.41 | 130459 ± 2052 |
| Llama | SJ | 100.00 ± 0.00 | 82.41 ± 0.00 | 68.75 ± 0.00 | 235.52 ± 8.46 | 133457 ± 2 |
| Llama | MJS | 100.00 ± 0.00 | 53.85 ± 0.18 | 43.75 ± 0.00 | 215.33 ± 5.13 | 69232 ± 10 |
| Llama | MCLR | 53.40 ± 0.00 | 22.84 ± 0.00 | 12.50 ± 0.00 | 174.31 ± 10.95 | 55823 ± 7 |

Note: Confidence bounds summarize variation across the four transport/constraint subconditions inside each backend × mode cell. They are descriptive stability bounds within the designed matrix rather than population confidence intervals.

### 5.2 Backend × mode interaction is the first-order effect

The central statistical result is that backend × mode interaction is not a minor nuance; it is a dominant determinant of the observed outcome vector. The factorial ANOVA summaries are shown below.

For routing accuracy, backend, mode, and the backend × mode interaction are all highly significant, with the interaction term itself showing a very large partial eta squared of 0.960. The same pattern holds for state retention and format compliance. Median latency also shows a significant backend × mode interaction, though with a more moderate effect size than correctness metrics. Token usage is dominated by mode, which is expected because token budget and output style are explicit design properties of the runtime profiles, while the backend × mode interaction for tokens is comparatively weak.

The important interpretive point is not merely that several p-values are small. It is that the interaction terms remain large even when judged by effect size. For RA, the backend × mode partial eta squared (0.960) is essentially on the same scale as the backend main effect (0.952) and only slightly below the mode main effect (0.993). For SR the interaction term again remains very large (0.939). In other words, interaction is not a residual curiosity after the main effects are explained away; it is one of the dominant sources of structured-routing variation in the entire design.

This matters methodologically because it invalidates one of the most common benchmarking shortcuts: averaging over backends and declaring a single package winner. In the present data, such averaging would erase the fact that the same burden-allocation move has very different consequences on Gemini, OpenAI, and Llama. The correct inference is therefore conditional rather than global: the utility of a runtime package depends on backend compatibility.

Table 3. Correctness-oriented factorial effects (Type-II ANOVA).

| Metric | Effect | F | p | Partial eta squared |
|---|---|---|---|---|
| RA% | Backend | 334.53 | <0.001 | 0.952 |
| RA% | Mode | 1519.07 | <0.001 | 0.993 |
| RA% | Constraint | 7.11 | 0.012 | 0.173 |
| RA% | Transport | 0.01 | 0.937 | 0.000 |
| RA% | Backend × mode | 136.52 | <0.001 | 0.960 |
| SR% | Backend | 242.28 | <0.001 | 0.934 |
| SR% | Mode | 365.63 | <0.001 | 0.970 |
| SR% | Constraint | 13.04 | <0.001 | 0.277 |
| SR% | Transport | 0.01 | 0.910 | 0.000 |
| SR% | Backend × mode | 87.56 | <0.001 | 0.939 |
| FC% | Backend | 203306.34 | <0.001 | 1.000 |
| FC% | Mode | 173852.49 | <0.001 | 1.000 |

| Metric | Effect | F | p | Partial eta squared |
|---|---|---|---|---|
| FC% | Constraint | 2.00 | 0.166 | 0.056 |
| FC% | Transport | 0.00 | 1.000 | 0.000 |
| FC% | Backend × mode | 174171.57 | <0.001 | 1.000 |

Note: Partial eta squared values are reported to emphasize substantive systems effects, not only nominal significance.

Table 4. Efficiency-oriented factorial effects (Type-II ANOVA).

Figure 4 condenses the ANOVA tables into an effect-size view and makes the hierarchy of influences easier to compare across correctness and efficiency metrics.

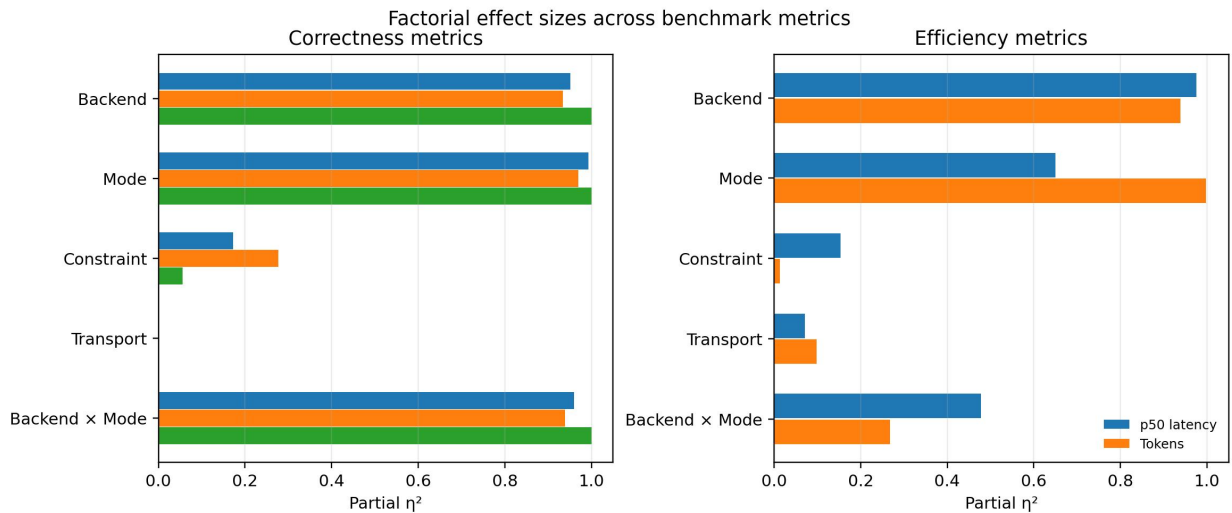

*Figure 4. Factorial effect-size overview across correctness and efficiency metrics.*

| Metric | Effect | F | p | Partial eta squared |
|---|---|---|---|---|
| p50 latency | Backend | 689.38 | <0.001 | 0.976 |
| p50 latency | Mode | 20.99 | <0.001 | 0.649 |
| p50 latency | Constraint | 6.15 | 0.018 | 0.153 |
| p50 latency | Transport | 2.62 | 0.115 | 0.072 |
| p50 latency | Backend × mode | 5.19 | <0.001 | 0.478 |
| Tokens | Backend | 258.28 | <0.001 | 0.938 |
| Tokens | Mode | 4929.32 | <0.001 | 0.998 |
| Tokens | Constraint | 0.47 | 0.496 | 0.014 |
| Tokens | Transport | 3.73 | 0.062 | 0.099 |
| Tokens | Backend × mode | 2.07 | 0.083 | 0.268 |

Note: For p50 latency, backend × mode remains significant with a moderate effect size; for tokens, the dominant main effect is mode.

### 5.2.1 Stability inside backend × mode cells

A second statistical question is whether the findings are being driven by noisy fluctuations inside each backend × mode cell rather than by genuine movement across the designed factors. The answer is largely no. For the correctness metrics, many backend × mode cells are either exactly invariant across the four subconditions or vary only negligibly. Gemini MJ and Gemini SJ, for example, remain fixed at 86.11% RA and 75.00% SR across their four constraint/transport variants. Llama SJ is similarly stable at 82.41%

RA and 68.75% SR. OpenAI MJ and OpenAI SJ remain above 85% RA despite the two transport and two constraint settings.

This within-cell stability is analytically useful. It means that the large interaction effect is not an artifact of high random variability within individual packages. Instead, the benchmark behaves as a designed systems experiment should: the dominant movements occur when the deployment profile actually changes. Where within-cell variability does appear, it is more visible in latency-sensitive quantities and in route-specific slices than in the aggregate RA/SR summaries.

A practical consequence follows. Engineers choosing among runtime packages should treat the backend–mode pair as the primary decision unit and the transport/constraint toggles as secondary refinements. The statistics support that ordering of priorities.

### 5.2.2 Targeted contrasts and cell-level confidence bounds

A stronger test of the compatibility claim comes from targeted within-backend contrasts. To avoid overstating population inference, the paper treats these contrasts as cell-level stability summaries over the four transport/constraint subconditions inside each backend x mode cell. Even under that conservative framing, the patterns are decisive.

Figure 2 visualizes the route-accuracy landscape behind the targeted contrasts and highlights the practical asymmetry between stable JSON packages and backend-sensitive compressed reconstruction.

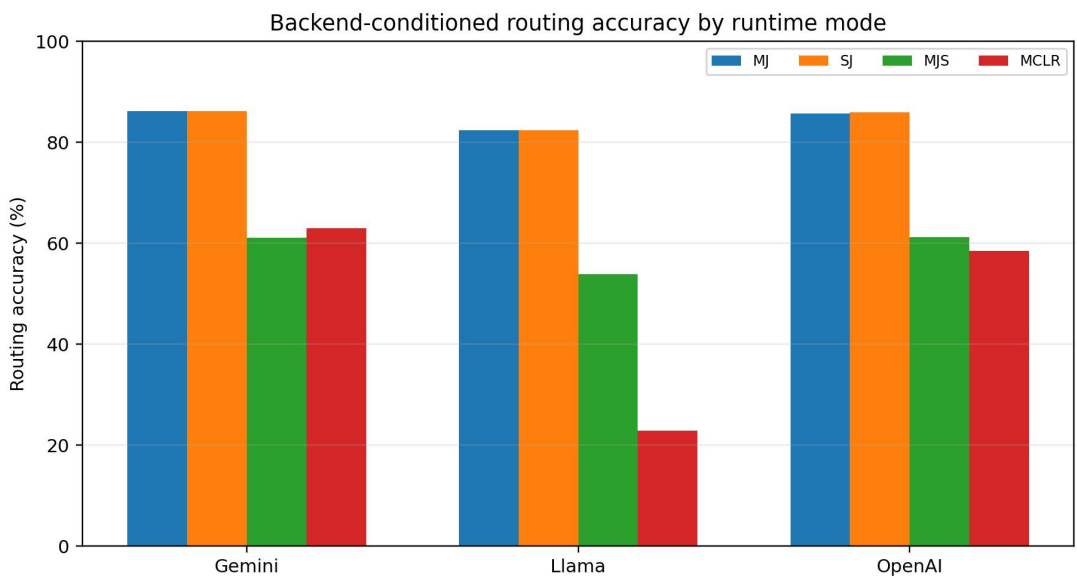

*Figure 2. Route-accuracy patterns by backend and runtime mode.*

On Gemini, MCLR is consistently faster and far cheaper than MJ/SJ, but the gain is purchased with a fixed 23.15-point drop in RA and a 6.25-point drop in SR relative to both JSON baselines. The latency advantage remains meaningful at about 139-148 ms median savings. On OpenAI, the trade-off is even sharper: MCLR improves median latency by about 380 ms and cuts tokens by roughly 75.8k relative to MJ/SJ, yet loses about 27 points of RA and 22 points of SR. On Llama, aggressive burden reallocation is not a mild degradation but a collapse: compared with SJ, MCLR loses 59.57 points of RA and 56.25 points of SR, even though it remains about 61 ms faster in median latency.

These targeted contrasts matter for two reasons. First, they show that the compatibility principle is not driven by averaging artifacts. Second, they identify an operational decision boundary: when correctness loss of 20-60 points accompanies the same efficiency move, compression cannot be justified by generic cost rhetoric and must instead be treated as a backend-specific deployment decision.

Table 5. Targeted MCLR contrasts against JSON baselines within each backend.

| Backend | Contrast | RA% | SR% | p50(ms) | Tokens |
|---|---|---|---|---|---|
| Gemini | MCLR vs MJ | -23.15 [-23.15, -23.15] | -6.25 [-6.25, -6.25] | -139.19 [-236.96, -44.16] | -80327 [-80327, -80327] |
| Gemini | MCLR vs SJ | -23.15 [-23.15, -23.15] | -6.25 [-6.25, -6.25] | -147.73 [-229.80, -67.81] | -80327 [-80327, -80327] |
| OpenAI | MCLR vs MJ | -27.16 [-30.55, -23.76] | -22.05 [-26.91, -17.19] | -382.19 [-486.38, -278.00] | -75774 [-81316, -70121] |
| OpenAI | MCLR vs SJ | -27.39 [-30.71, -24.15] | -22.05 [-26.91, -17.19] | -379.98 [-495.28, -266.92] | -75780 [-81351, -70154] |
| Llama | MCLR vs MJ | -59.49 [-59.57, -59.34] | -56.25 [-56.25, -56.25] | -57.58 [-67.67, -47.73] | -74636 [-76444, -72825] |
| Llama | MCLR vs SJ | -59.57 [-59.57, -59.57] | -56.25 [-56.25, -56.25] | -61.21 [-73.17, -49.49] | -77634 [-77640, -77628] |

Note: Bracketed values denote descriptive 95% bootstrap intervals over the four subconditions inside each backend × mode cell. Negative values indicate that MCLR underperforms the JSON baseline on correctness or uses fewer resources on efficiency metrics.

5.2.3 Bounded decomposition of the MCLR package

The benchmark does not contain a dedicated CLR-high-budget rerun or a compact-no-reconstruction rerun, so a full intervention-level ablation of MCLR remains future work. That limitation should be acknowledged directly. However, the existing factorial design already separates one potentially confounding component: output-budget relaxation inside direct JSON emission. Related function-calling work likewise shows that packaging choices, reasoning traces, and argument realization interact rather than decomposing cleanly into a single budget effect .

Across Gemini and OpenAI, the MJ-to-SJ contrast is essentially flat on correctness. Both backends preserve the same routing accuracy and state retention under the low-budget and high-budget JSON baselines. Llama changes only marginally, by about +0.08 points in RA with no SR improvement. This matters because it narrows the bundle criticism. The large correctness losses seen under MCLR cannot reasonably be attributed to budget changes alone.

What remains bundled is the move from direct JSON emission to compact semantic coding plus local deterministic reconstruction. The current evidence should therefore be read as a bounded decomposition rather than a full causal ablation. Even so, it already supports a stronger methodological statement than earlier versions of the paper: within direct JSON emission, budget relaxation by itself does not generate the 20–60 point correctness losses observed for MCLR. The dominant risk is therefore tied to compatibility of compact coding and reconstruction with the chosen backend, not to token budget alone.

## 5.3 Correctness profiles across backends

### Gemini

Gemini favors direct JSON generation. MJ and SJ are tied on RA (86.11%) and SR (75.00%), with perfect FC. MCLR preserves FC but drops to 62.96% RA and 68.75% SR. MJS performs even worse on correctness, falling to 61.03% RA and 56.08% SR. The implication is that for Gemini, reducing structural burden and/or introducing streaming does not improve the control decision itself. The backend appears most reliable when asked to emit the final structured record directly.

**OpenAI**

OpenAI shows a similar but not identical pattern. MJ and SJ again remain the strongest correctness profiles, at 85.65–85.88% RA and 71.88% SR, with 100% FC throughout. MCLR is faster and much cheaper in tokens, but RA drops to 58.49% and SR to 49.82%. MJS lands between the high-fidelity JSON baselines and MCLR in cost, but its RA is only 61.11%. Thus, OpenAI also supports the compatibility principle: compression helps efficiency, but high-fidelity structured emission remains the safer choice when correctness and statefulness dominate.

**Llama**

Llama exhibits the sharpest incompatibility. MJ and SJ still preserve reasonably strong correctness, at about 82.3–82.4% RA and 68.75% SR, though MJ loses some FC. MJS degrades correctness substantially, and MCLR collapses further still, with 53.40% FC, 22.84% RA, and 12.50% SR. Yet Llama is also the fastest backend in absolute latency. This creates a compelling systems trade-off: the fastest backend is not necessarily the most compatible backend for aggressive burden reallocation.

## 5.4 Efficiency gains are real, but backend-conditioned

The second major result concerns efficiency. Across all backends, MCLR is the strongest efficiency profile. It uses the fewest tokens and often the lowest median latency. Averaged over all backends, mode-level means show that MCLR reduces token usage to about 47.2k, compared with 124.1k for MJ and 125.1k for SJ. Median latency also falls to about 616.64 ms, compared with about 810–813 ms for MJ, MJS, and SJ. These are not trivial savings. Related prompt-compression and structured-generation studies similarly report that representation choices can materially change token cost and end-to-end responsiveness (Jiang, Wu, Lin, Yang, & Qiu, 2023).

However, the study does not support the stronger claim that efficiency-preserving correctness follows automatically. Compression helps when the backend can carry the semantics reliably through the compact intermediate representation. Where that condition is not met, the efficiency gain is purchased by a loss of route correctness, state retention, or even format compliance. This is precisely what Proposition 2 anticipated.

The strongest methodological interpretation is therefore not "MCLR is best" but "burden reallocation changes the position of a deployment on a correctness-cost-latency frontier." That frontier is backend-conditioned.

Figure 3 makes the compatibility principle concrete by showing that the relevant comparison is a backend-conditioned frontier rather than a universal winner across all runtime packages.

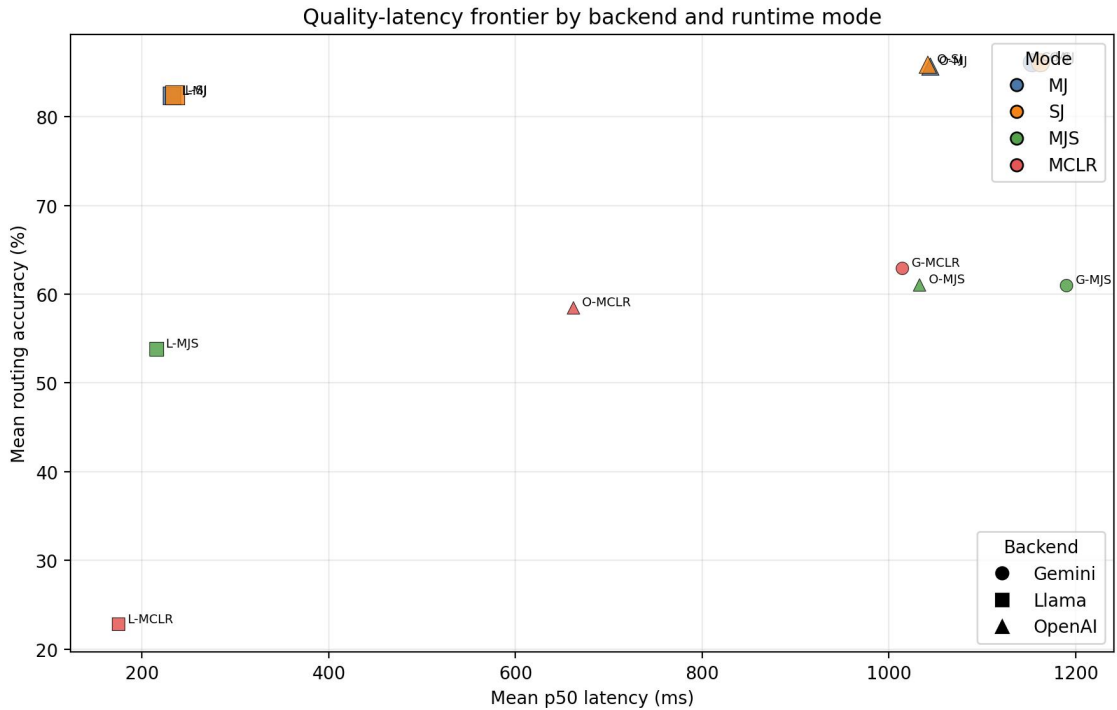

Figure 3. Backend-conditioned correctness–efficiency frontier.

### 5.4.1 Workflow lower-bound completion (WLC)

The workflow relevance of the routing metrics becomes sharper when they are combined into the derived WLC proxy. Because WLC is a lower bound on the joint event of format-validity, correct routing, and state-safe control, it estimates the minimum rate at which the router can hand downstream orchestration a control artifact that is both usable and correctly directed. This moves the discussion one step closer to expert-system operational value without pretending to measure the final downstream answer quality.

Table 6. Workflow lower-bound completion (WLC) by backend and runtime mode.

Figure 5 gives a compact visual summary of WLC and highlights that workflow-safe completion remains strongly backend-conditioned even after a common lower-bound definition is applied.

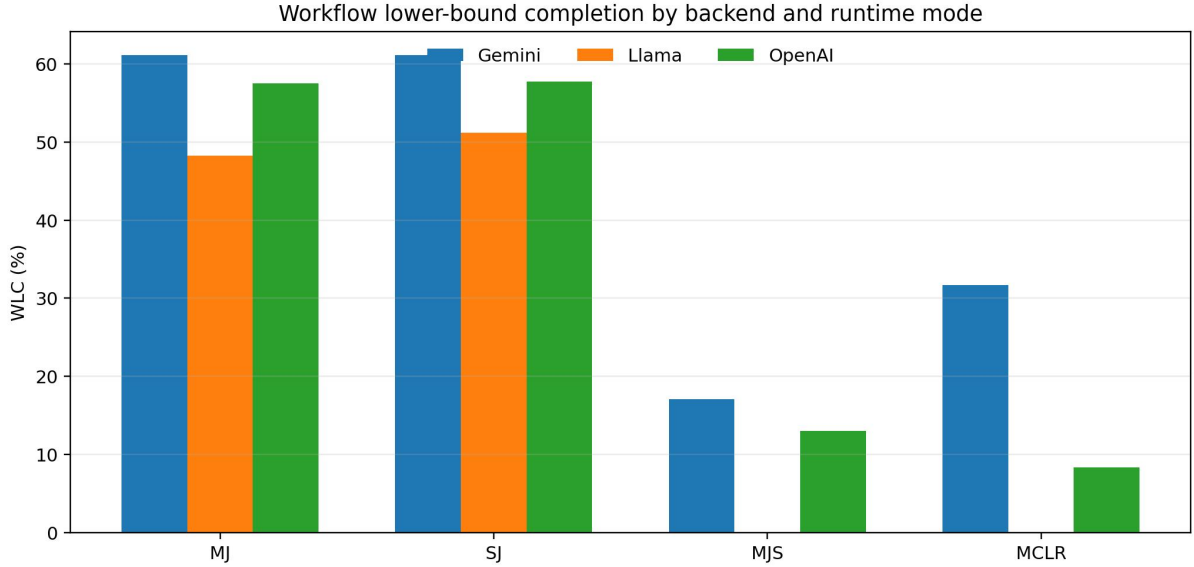

*Figure 5. Workflow lower-bound completion (WLC) by backend and runtime mode.*

| Backend | Mode | WLC% |
|---------|------|------|
| Gemini | MJ | 61.11 |
| Gemini | SJ | 61.11 |
| Gemini | MJS | 17.03 |
| Gemini | MCLR | 31.71 |
| OpenAI | MJ | 57.53 |
| OpenAI | SJ | 57.76 |
| OpenAI | MJS | 13.02 |
| OpenAI | MCLR | 8.31 |
| Llama | MJ | 48.22 |
| Llama | SJ | 51.16 |
| Llama | MJS | 0.00 |
| Llama | MCLR | 0.00 |

The WLC view clarifies the operational stakes. On Gemini, direct JSON packages (MJ/SJ) sustain a lower-bound workflow completion of 61.11%, whereas MCLR falls to 31.71% and MJS to 17.03%. On OpenAI, MJ/SJ remain near 57.5%, while MCLR falls to 8.31%. On Llama, the lower-bound workflow completion remains 48.22–51.16% for MJ/SJ but collapses to 0.00% for both MJS and MCLR. This is exactly the kind of deployment signal that an expert-system operator needs: a package can still look efficient on aggregate latency and token cost while offering almost no guaranteed floor on usable downstream control completions.

### 5.5 Streaming and constraint settings are secondary effects

Figure 6 shows that tail-latency amplification varies by backend and mode, but the resulting pattern is still smaller than the dominant backend × mode interaction observed for correctness.

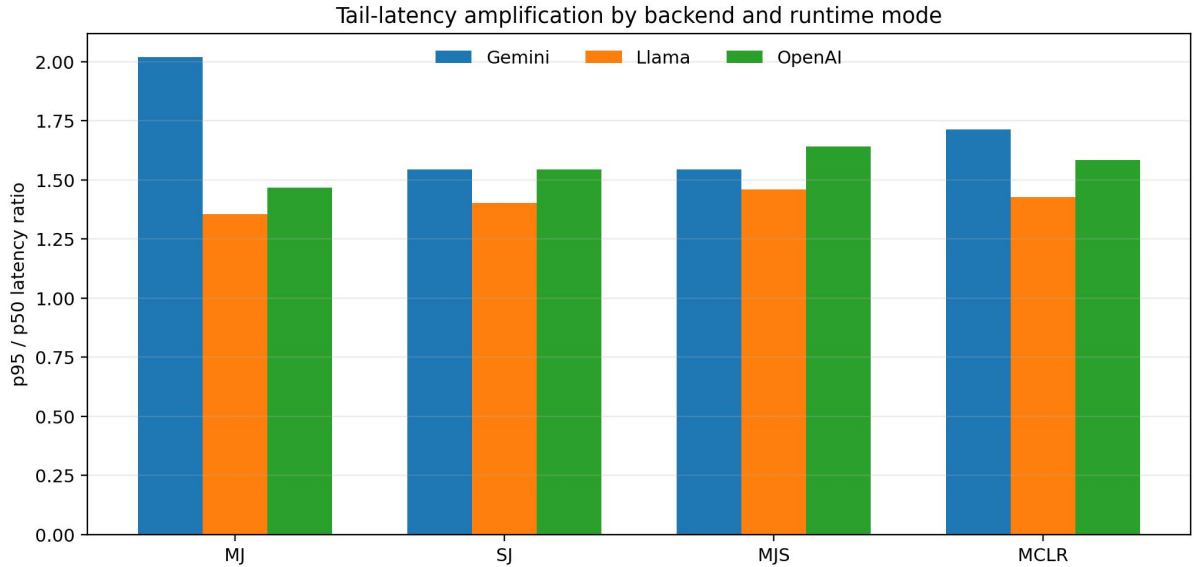

*Figure 6. Tail-latency amplification (p95/p50) by backend and runtime mode.*

A third result concerns transport and constraint. Descriptively, transport has only small effects on the core correctness metrics. The ANOVA confirms this. For RA, the transport term is negligible ($p = 0.9365$). For SR and FC it is similarly negligible. Even for median latency, transport is not significant at the conventional 0.05 threshold in this combo-level model ($p = 0.1149$), though some specific combinations do display transport-related latency differences.

Constraint has modest but statistically detectable effects on RA, SR, and p50 latency. Unlimited settings improve mean RA slightly (69.70% vs. 68.43%) and reduce latency somewhat (736.55 ms vs. 789.33 ms). However, the effect sizes are much smaller than those for backend, mode, and backend × mode interaction. This means that relaxing budget or reasoning profile can help at the margin, but it is not the dominant lever for structured routing quality. Runtime packaging matters more.

This pattern supports Proposition 3. For control tasks, transport semantics contribute far less than one might assume from user-facing chat applications. Since the full control record must be available before downstream execution can proceed, streaming provides limited operational value.

### 5.6 Failure taxonomy

Figure 7 decomposes attrition from structure validity to routing correctness and state retention, making the distinct loss mechanisms visible at the backend–mode level.

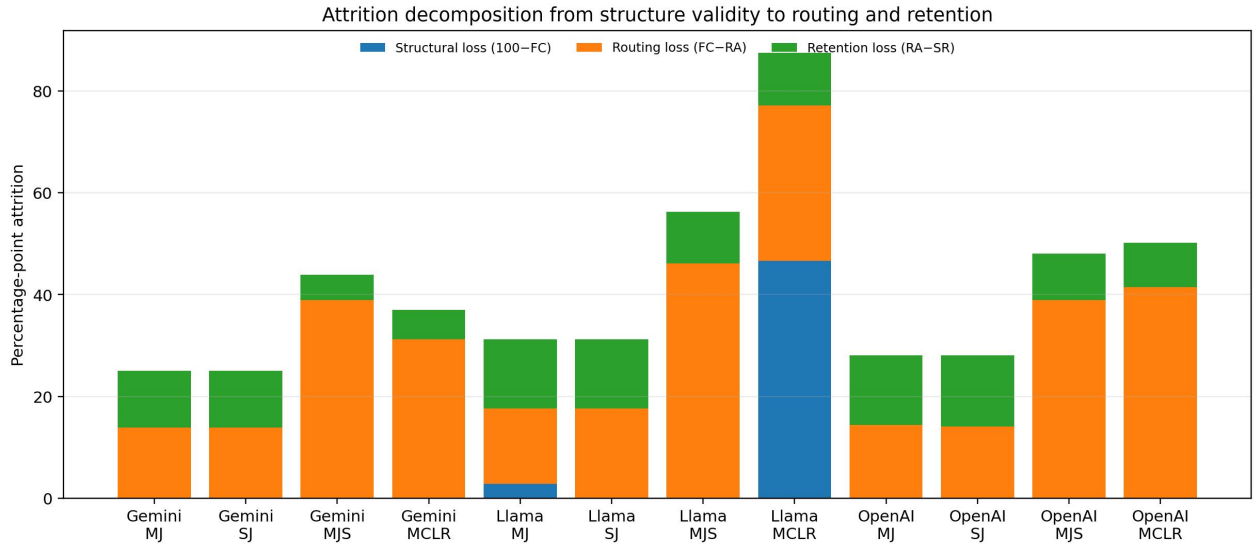

*Figure 7. Attrition decomposition from structure validity to routing correctness and state retention.*

The benchmark also records a concise failure taxonomy. Across all 15,552 requests, 14,910 complete without failure, while the remaining failures are dominated by JSON parse errors (523) and schema-invalid outputs (100), with only 18 HTTP 400 errors and 1 rate-limit event. This composition is informative.

First, the failure profile is overwhelmingly structural rather than infrastructural. The benchmark is not primarily being limited by transport failures or API instability (Zhu, Yan, Shi, Yin, & Sha, 2024). It is being limited by the system's ability to produce machine-actionable structured control artifacts. Second, the presence of both parse errors and schema-invalid outputs supports the burden-allocation perspective. These are exactly the kinds of failure modes one would expect when structure realization is part of the workload.

Failure taxonomy is therefore not merely a debugging appendix; it is evidence for mechanism. If compressed local reconstruction eliminates certain surface formatting burdens but introduces semantic brittleness in the compact code, one would expect some structural errors to decline and others to reappear as correctness loss rather than parse failure. That is broadly consistent with the observed pattern, especially on Llama under MCLR.

5.7 Route-level fragility and protected specialist routes

Figure 8 makes route-level fragility explicit and shows that the most severe degradation is concentrated in specialist routes rather than broad conversational routes.

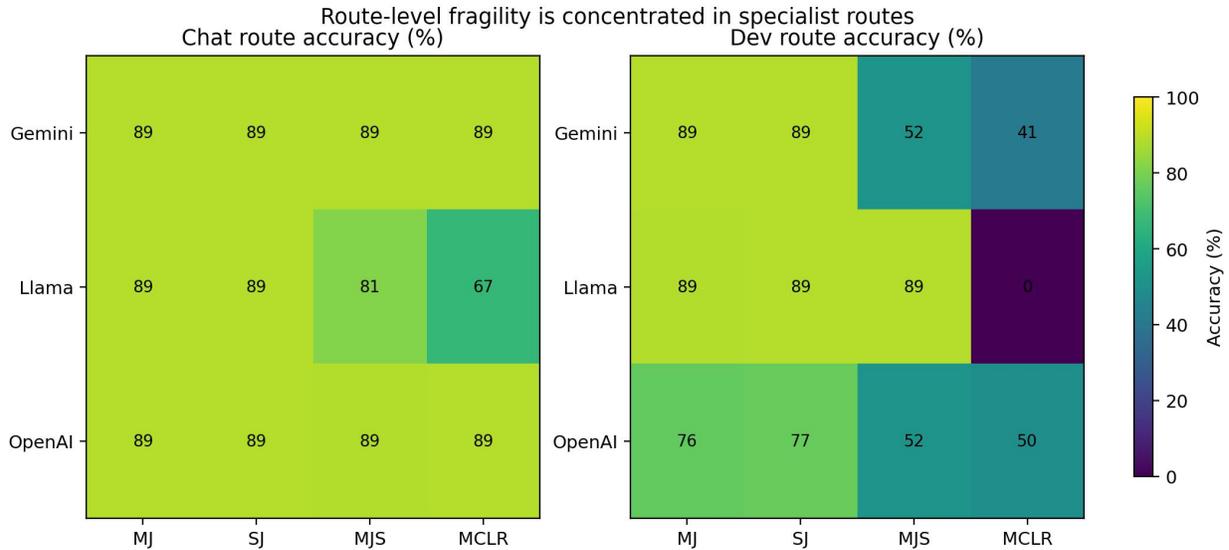

*Figure 8. Route-level fragility heatmaps for chat and developer routes.*

The benchmark summary also reports route-specific slices, including chat and dev accuracy. These route-level metrics matter because aggregate RA can hide selective fragility. In the present benchmark, chat routing is comparatively robust across most packages. Gemini remains at 88.89% chat accuracy across all four modes. OpenAI is similarly stable near 88.89%, with only a slight decrease for MJS. Even Llama retains 88.89% chat accuracy under MJ and SJ and only drops sharply under MCLR. Related robustness and agent benchmark work similarly warns that aggregate success can hide instability in specific control paths or subskills (Debenedetti et al., 2024).

Developer-oriented routing is a different story. The dev route is where packaging fragility becomes highly visible. On Gemini, dev accuracy is 88.89% under MJ and SJ but falls to 40.74% under MCLR and 51.85% under MJS. On OpenAI, dev accuracy declines from roughly 75.93–76.85% under MJ/SJ to 49.69% under MCLR and 51.85% under MJS. Llama shows the strongest asymmetry of all: dev accuracy remains about 88.6–88.9% under MJ/SJ, stays high under MJS, but collapses to 0.00% under MCLR.

These route-level contrasts strengthen the theoretical interpretation in two ways. First, they show that burden reallocation is not just a global softening of accuracy; it selectively reshapes the representational distinctions the router can preserve. Second, they provide a deployment warning that aggregate RA alone would understate: a package can remain acceptable on high-frequency conversational routing while becoming operationally unsafe on a business-critical specialist route.

This selective fragility is one reason the methodology is better viewed as a structured-control evaluation protocol than as a generic text benchmark. For front-door expert-system dispatch, the operational question is often whether the expensive or irreversible routes remain protected. A packaging choice that sacrifices those routes in exchange for token savings may be undesirable even when its aggregate averages appear tolerable.

### 5.8 Summary of empirical support for the propositions

The empirical support for the framework can now be summarized succinctly.

- Proposition 1 (interaction proposition): supported. Backend × mode interaction is statistically strong for RA, SR, FC, and p50 latency.
- Proposition 2 (recoverable-structure proposition): conditionally supported. Compressed local reconstruction delivers strong efficiency gains, but correctness preservation is backend-dependent and not universal.
- Proposition 3 (actionable-latency proposition): supported. Transport has negligible effect on core correctness and only a limited effect on full-response latency, suggesting that streaming is not a major lever for compact structured control.

## 6. Discussion

### 6.1 From benchmark to methodology

The main value of this study is not a leaderboard. It is a way of thinking about structured routing as a systems design problem. Traditional benchmark narratives often ask which mode or model is "best." That framing is too coarse for expert-systems deployment because it suppresses the distinction between semantic inference and structure realization. By introducing burden allocation as an explicit design dimension, this paper makes that distinction observable and testable.

The result is a methodology with four steps:

1. Define a compact control ontology. Specify the fields that make the routing artifact operationally actionable.
2. Enumerate burden-allocation dimensions. Decide how much serialization burden remains in the model loop, whether transport is streamed, and where final structure is realized.
3. Measure both correctness and operational cost. Do not collapse evaluation into route accuracy alone.
4. Analyze interaction, not just ranking. Backend compatibility is itself an empirical object.

This methodological framing is one of the strongest reasons the work can contribute to an ESWA audience. It yields deployment rules and evaluation practice, not merely a one-off benchmark result. Recent surveys on tool learning, agent security, domain specialization, and broader LLM deployment similarly argue that practical behavior emerges from the interaction between models, tools, interfaces, and application constraints rather than from model quality alone (Zeeshan et al., 2025).

### 6.2 Theoretical meaning of burden allocation

The theory proposed here is deliberately lightweight, but it is not empty. Its core claim is that runtime packaging redistributes cognitive and structural work across the model and the surrounding software. In a direct-JSON mode, the model must solve both the semantic task ("which route?") and the interface task ("how do I serialize that choice into the target schema?"). In a compressed local-reconstruction mode, the model solves a smaller emission task while deterministic software expands the result into the final record. This reallocation can reduce burden, but only if the compressed code remains semantically reliable.

That perspective explains why one should not expect monotonic improvements from "making the output simpler." Simpler for the interface is not always simpler for the model. A compact code can reduce serialization overhead while also removing lexical anchors or disambiguating redundancy that help a particular backend preserve route semantics. In other words, burden allocation changes both cost and representational geometry (Voboril, Ramaswamy, & Szeider, 2025).

This interpretation also clarifies why the framework is best described as a compatibility theory rather than a universal optimization theory. It explains *when* a packaging move is likely to help: when the structure being offloaded is recoverable and low-entropy, and when the backend can still stably emit the compressed semantic kernel.

### 6.3 Why the current abstraction is useful even without a closed-form predictor

One reasonable criticism is that the framework does not yet provide a closed-form predictor of the optimal runtime package. That criticism is correct. However, practical systems theory often advances first by identifying state variables, dominant interactions, and stable design rules before it reaches fully predictive form. In that sense, the burden-allocation framework already does meaningful theoretical work. It identifies the relevant dimensions, proposes testable propositions, explains why backend-independent optimality should not be assumed, and provides an experimental protocol through which compatibility can be assessed.

For an applied expert-systems audience, that is already valuable. Many engineering failures occur not because no formal predictor exists, but because a design variable has not yet been recognized as a variable at all. Runtime packaging in structured routing is one such variable.

### 6.4 Deployment implications

Three deployment rules follow from the results.

Rule 1: Preserve direct structured emission when correctness dominates and the backend supports it well.
On Gemini and OpenAI, MJ and SJ preserve correctness most consistently. If routing errors or state failures are expensive, high-fidelity JSON emission should remain the default.

Rule 2: Use compressed local reconstruction when efficiency dominates and compatibility has been verified.
MCLR is the strongest efficiency profile, but only a safe choice where its semantic compression does not destabilize route choice. It should therefore be deployed only after backend-specific validation rather than assumed to be generally safe.

Rule 3: Do not overvalue streaming in compact control tasks.
Since downstream execution cannot proceed until the full control record is available, streaming contributes far less than in user-facing free-form chat. For compact structured control, actionable latency matters more than token-by-token partial display.

These rules are exactly the sort of practical guidance that expert-systems journals value: they transform empirical findings into design decisions. They also align with the broader systems literature showing that orchestration quality, route selection, and workflow integration are now central determinants of practical value in LLM-enabled applications.

6.4.1 Backend-conditioned selection procedure

Step 1: Protect the irreversible or specialist routes first. If wrong dispatch on developer, document, or tool-activation branches is operationally expensive, begin from the direct-JSON package with the strongest route-specific safety and WLC, not from the cheapest token profile.

Step 2: Screen compact packages against a workflow floor, not only against mean latency. A compressed package is admissible only if its RA, SR, and WLC remain within the deployment tolerance of the host workflow.

Step 3: Use streaming only as a secondary tie-breaker. When downstream execution cannot start from partial tokens, transport should be selected after correctness and workflow-safe completion are already acceptable.

Step 4: Reject one-size-fits-all deployment assumptions. If a backend shows specialist-route fragility or WLC collapse under compact coding, that incompatibility should be treated as a backend-local property and not averaged away by global rankings.

## 6.5 Bridging routing quality to workflow loss

A remaining challenge for any routing study is to explain why routing metrics alone should matter to expert-systems readers who ultimately care about downstream workflow performance. The present reconstruction now makes that bridge partially empirical through WLC, a conservative lower-bound estimate of workflow-safe control completion derived from FC, RA, and SR. The answer is that a front-door router does not need to solve the entire downstream task to create or prevent workflow value; it only needs to issue a usable control record that sends the request into the correct branch with enough state continuity for execution to proceed.

For that reason, FC, RA, and SR are not merely proxy scores borrowed from benchmark culture. They correspond to three distinct categories of workflow loss: interruption loss (the controller emits unusable structure), branch loss (the controller selects the wrong subsystem), and continuity loss (the controller fails to preserve the state required for the next action). In production systems these losses translate into user-visible friction, wasted tokens, redundant tool calls, and delayed task completion. The present study does not yet assign a calibrated monetary value to each loss category, but it does establish the structural preconditions for such valuation.

This bridge also clarifies why actionable latency is the correct temporal measure for compact control. When no partial token stream can trigger downstream execution, only the arrival of the complete control artifact reduces workflow delay. The operational object is therefore not conversational fluency but decision availability.

## 6.6 External validity and scope conditions

A second criticism concerns task-space narrowness. The benchmark studies a four-route ontology with a compact prompt pool and does not directly measure downstream final task success. This limitation is real and must be stated clearly. However, it does not invalidate the contribution; it bounds it.

The methodology is best suited to small-schema, enumerable-control, front-door routing tasks. Examples include enterprise copilot dispatch, document-versus-chat separation, workflow triage, tool preselection, and modular assistant orchestration (Yeung et al., 2025). These settings share three characteristics: (i) the control record is compact, (ii) misroutes have operational cost, and (iii) the router must emit machine-readable structure reliably.

By contrast, the methodology should not be overclaimed for very large or open-ended route ontologies, heavy semantic decomposition tasks, or settings where the final downstream answer quality dominates all routing errors. In those settings, broader end-to-end task evaluation would be essential.

### 6.7 Statistical hardness and what has improved

Another criticism of earlier versions of the study was that the argument sounded plausible but statistically soft. The present reconstruction addresses that problem by explicitly testing designed effects. The significance and effect-size results show that backend and mode are not small perturbations; they explain large portions of the systematic variance in RA, SR, and FC. The interaction term is not merely significant but substantively large. This matters because it transforms the paper's main claim from a verbal impression into an experimentally supported systems statement.

At the same time, one should not overstate what ANOVA can do here. The unit of analysis is the combo-level deployment profile, not an open-world stream of naturally sampled production requests. The statistical conclusions therefore quantify designed-factor effects within the experimental matrix. They do not prove universal generalization beyond the studied task class.

### 6.8 Threats to validity

Several threats to validity remain.

Task-space limitation. The route ontology is intentionally compact, with four route labels. This keeps the experiment interpretable but limits immediate transfer to much larger routing spaces.

Prompt-pool limitation. The prompt pool is diverse enough to expose failure regimes, but still much smaller than real production traffic. Additional prompt families may reveal new fragilities.

No direct downstream task-success metric. The study measures routing quality, structure validity, and state retention, not the final answer quality of the selected downstream subsystem. Future work should connect routing errors to actual downstream outcome loss.

Intervention bundling. MCLR is a package intervention rather than a one-factor ablation. It combines compact emission, lower token budget, and local reconstruction. This is appropriate for deployment comparison, but it means the marginal effect of each micro-design choice is not separately identified.

Lightweight theory. The burden-allocation framework identifies variables and interactions but does not yet yield a closed-form optimality predictor. The theory should therefore be understood as an explanatory and methodological abstraction rather than a completed predictive law.

These limitations are serious, but they are also productive. They define a concrete future-work agenda rather than undermining the present contribution.

### 7. Conclusion

This paper has reconstructed structured LLM routing as a runtime burden-allocation problem for agentic expert systems. The evidence comes from a full-factorial benchmark spanning 48 deployment combinations and 15,552 routed requests. The key result is not that one mode wins everywhere. It is that runtime packaging materially changes structured-control behavior and that the direction of the change depends on backend family.

That claim is supported both descriptively and statistically. Backend × mode interaction is a first-order effect for routing accuracy, state retention, format compliance, and median latency. High-fidelity JSON modes remain the safest correctness profiles on Gemini and OpenAI. Compressed local reconstruction is the strongest efficiency profile, but its benefits are backend-conditioned rather than universal. Streaming contributes little to control correctness and only secondary value to actionable latency.

The broader implication is methodological. Expert-system designers should evaluate front-door LLM routing as a structured-control component with explicit burden-allocation choices, not as a miniature chat task and not as a prompt tweak. Once runtime packaging is treated as a first-class design variable, the relevant engineering question becomes compatibility: which burden-allocation profile places the chosen backend on the best correctness-cost-latency frontier for the target application?

Future work should extend this methodology to larger route ontologies, direct downstream task-success metrics, and more granular decomposition of bundled package interventions. Even in its present form, however, the framework contributes a practical and theoretically grounded way to reason about structured routing in expert systems.

**Data availability statement**

An anonymized supplementary package accompanies this submission and contains the benchmark matrix, combo-level summaries, route-slice aggregates, ANOVA outputs, and the deterministic local reconstruction logic used in the study. Prompt templates and schema definitions are included in sanitized form. Provider credentials, endpoint secrets, and environment-specific identifiers have been removed.

References

Alon, U., Clark, P., Dziri, N., Gao, L., Gupta, P., Gupta, S., . . . Yazdanbakhsh, A. (2023). *Self-Refine: Iterative Refinement with Self-Feedback*. Paper presented at the Advances in Neural Information Processing Systems 36. https://doi.org/10.52202/075280-2019

Ayala, O., & Bechard, P. (2024). *Reducing hallucination in structured outputs via Retrieval-Augmented Generation*. Paper presented at the Proceedings of the 2024 Conference of the North American Chapter of the Association for Computational Linguistics: Human Language Technologies (Volume 6: Industry Track).

Bosma, M., Chi, E., Ichter, B., Le, Q. V., Schuurmans, D., Wang, X., . . . Zhou, D. (2022). *Chain-Of-Thought Prompting Elicits Reasoning in Large Language Models*. Paper presented at the Advances in Neural Information Processing Systems 35. https://doi.org/10.52202/068431-1800

Cancedda, N., Dessi, R., Dwivedi-Yu, J., Hambro, E., Lomeli, M., Raileanu, R., . . . Zettlemoyer, L. (2023). *Toolformer: Language Models Can Teach Themselves to Use Tools*. Paper presented at the Advances in Neural Information Processing Systems 36. https://doi.org/10.52202/075280-2997

Chen, Z.-Y., Shen, S., Shen, G., Zhi, G., Chen, X., & Lin, Y. (2024). *Towards tool use alignment of large language models*. Paper presented at the Proceedings of the 2024 Conference on Empirical Methods in Natural Language Processing.

Cheng, S., Zhuang, Z., Xu, Y., Yang, F., Zhang, C., Qin, X., . . . Zhang, D. (2024). *Call me when necessary: Llms can efficiently and faithfully reason over structured environments*. Paper presented at the Findings of the Association for Computational Linguistics: ACL 2024.

Daiber, J., Maricato, V., Sinha, A., & Rabinovich, A. (2025). *DispatchQA: A Benchmark for Small Function Calling Language Models in E-Commerce Applications*. Paper presented at the Proceedings of the 2025 Conference on Empirical Methods in Natural Language Processing: Industry Track. https://doi.org/10.18653/v1/2025.emnlp-industry.154

Debenedetti, E., Zhang, J., Balunovic, M., Beurer-Kellner, L., Fischer, M., & Tramèr, F. J. A. i. N. I. P. S. (2024). Agentdojo: A dynamic environment to evaluate prompt injection attacks and defenses for llm agents. *37*, 82895-82920.

Erdogan, L. E., Lee, N., Jha, S., Kim, S., Tabrizi, R., Moon, S., . . . Gholami, A. (2024). *Tinyagent: Function calling at the edge.* Paper presented at the Proceedings of the 2024 Conference on Empirical Methods in Natural Language Processing: System Demonstrations.

Greshake, K., Abdelnabi, S., Mishra, S., Endres, C., Holz, T., & Fritz, M. (2023). *Not What You've Signed Up For: Compromising Real-World LLM-Integrated Applications with Indirect Prompt Injection*. Paper presented at the Proceedings of the 16th ACM Workshop on Artificial Intelligence and Security. https://doi.org/10.1145/3605764.3623985

Guo, T., Chen, X., & Wang, Y. (2024). Large Language Model based Multi-Agents: A Survey of Progress and Challenges. *arXiv preprint*. doi:10.48550/arxiv.2402.01680

He, F., Zhu, T., Ye, D., Liu, B., Zhou, W., & Yu, P. S. J. A. C. S. (2025). The emerged security and privacy of llm agent: A survey with case studies. *58*(6), 1-36.

Jiang, H., Wu, Q., Lin, C.-Y., Yang, Y., & Qiu, L. (2023). *LLMLingua: Compressing Prompts for Accelerated Inference of Large Language Models*. Paper presented at the Proceedings of the 2023 Conference on Empirical Methods in Natural Language Processing. https://doi.org/10.18653/v1/2023.emnlp-main.825

Jimenez, C. E., Yang, J., Wettig, A., Yao, S., Pei, K., Press, O., & Narasimhan, K. J. a. p. a. (2023). Swe-bench: Can language models resolve real-world github issues?

Kernan Freire, S., Wang, C., Foosherian, M., Wellsandt, S., Ruiz-Arenas, S., & Niforatos, E. J. F. i. A. i. (2024). Knowledge sharing in manufacturing using LLM-powered tools: user study and model benchmarking. *7*, 1293084.

Khattab, O., Singhvi, A., Maheshwari, P., Zhang, Z., Santhanam, K., Vardhamanan, S., . . . Moazam, H. J. a. p. a. (2023). Dspy: Compiling declarative language model calls into self-improving pipelines.

Li, D., Lu, W., Shen, Y., Song, K., Tan, X., & Zhuang, Y. (2023). *HuggingGPT: Solving AI Tasks with ChatGPT and its Friends in Hugging Face*. Paper presented at the Advances in Neural Information Processing Systems 36. https://doi.org/10.52202/075280-1657

Ling, C., Zhao, X., Lu, J., Deng, C., Zheng, C., Wang, J., . . . Li, Y. (2025). Domain Specialization as the Key to Make Large Language Models Disruptive: A Comprehensive Survey. *ACM Computing Surveys*. doi:10.1145/3764579

Liu, X., Yu, H., Zhang, H., Xu, Y., Lei, X., Lai, H., . . . Yang, K. J. a. p. a. (2023). Agentbench: Evaluating llms as agents.

Opsahl-Ong, K., Ryan, M. J., Purtell, J., Broman, D., Potts, C., Zaharia, M., & Khattab, O. (2024). *Optimizing instructions and demonstrations for multi-stage language model programs.* Paper presented at the Proceedings of the 2024 Conference on Empirical Methods in Natural Language Processing.

Qin, Y., Hu, S., Lin, Y., Chen, W., Ding, N., Cui, G., . . . Xiao, C. J. A. C. S. (2024). Tool learning with foundation models. *57*(4), 1-40.

Romain, C., & Sarath, S. (2024). Reasoning in Large Language Models: A Geometric Perspective. doi:10.48550/arxiv.2407.02678

Romero, O. J., Zimmerman, J., Steinfeld, A., & Tomasic, A. (2023). *Synergistic integration of large language models and cognitive architectures for robust ai: An exploratory analysis.* Paper presented at the Proceedings of the AAAI Symposium Series.

Saha, A., Mandal, L., Ganesan, B., Ghosh, S., Sindhgatta, R., Eberhardt, C., . . . Mehta, S. (2024). *Sequential api function calling using graphql schema.* Paper presented at the Proceedings of the 2024 Conference on Empirical Methods in Natural Language Processing.

Shen, Z., Wang, D. Y.-B., Mishra, S. S., Xu, Z., Teng, Y., & Ding, H. (2025). *SLOT: Structuring the Output of Large Language Models*. Paper presented at the Proceedings of the 2025 Conference on Empirical Methods in Natural Language Processing: Industry Track. https://doi.org/10.18653/v1/2025.emnlp-industry.32

Sienicki, K. (2024). *RouterBench: A benchmark for multi-LLM routing systems*. https://doi.org/10.20944/preprints202411.2377.v1

Tam, Z. R., Wu, C.-K., Tsai, Y.-L., Lin, C.-Y., Lee, H.-y., & Chen, Y.-N. J. a. p. a. (2024). Let me speak freely? a study on the impact of format restrictions on performance of large language models.

Tang, X., Zong, Y., Phang, J., Zhao, Y., Zhou, W., Cohan, A., & Gerstein, M. (2024). *Struc-Bench: Are Large Language Models Good at Generating Complex Structured Tabular Data?* Paper presented at the Proceedings of the 2024 Conference of the North American Chapter of the Association for Computational Linguistics: Human Language Technologies (Volume 2: Short Papers). https://doi.org/10.18653/v1/2024.naacl-short.2

Tavanaei, A., Koo, K. K., Ceker, H., Jiang, S., Li, Q., Han, J., & Bouyarmane, K. (2024). *Structured object language modeling (SO-LM): Native structured objects generation conforming to complex schemas with self-supervised denoising.* Paper presented at the Proceedings of the 2024 Conference on Empirical Methods in Natural Language Processing: Industry Track.

Voboril, F., Ramaswamy, V. P., & Szeider, S. (2025). *StreamLLM: Enhancing Constraint Programming with Large Language Model-Generated Streamliners.* Paper presented at the 2025 IEEE/ACM 1st International Workshop on Neuro-Symbolic Software Engineering (NSE).

Wallace, E., Xiao, K., Leike, R., Weng, L., Heidecke, J., & Beutel, A. J. a. p. a. (2024). The instruction hierarchy: Training llms to prioritize privileged instructions.

Wu, Y., Wang, W., & Zhang, J. (2023). *AutoGen: Enabling next-gen LLM applications via multi-agent conversation*.

Yao, S., Zhao, J., Yu, D., Du, N., Shafran, I., Narasimhan, K. R., & Cao, Y. (2022). *React: Synergizing reasoning and acting in language models.* Paper presented at the The eleventh international conference on learning representations.

Yeung, C., Yu, J., Cheung, K. C., Wong, T. W., Chan, C. M., Wong, K. C., & Fujii, K. J. a. p. a. (2025). A zero-shot LLM framework for automatic assignment grading in higher education.

Yoran, O., Amouyal, S. J., Malaviya, C., Bogin, B., Press, O., & Berant, J. (2024). *Assistantbench: Can web agents solve realistic and time-consuming tasks?* Paper presented at the Proceedings of the 2024 Conference on Empirical Methods in Natural Language Processing.

Yue, M. (2024). *Large Language Model Cascades with Mixture of Thought Representations for Cost-Efficient Reasoning*. Paper presented at the International Conference on Learning Representations (ICLR).

Zeeshan, H. M. A., Umer, M., Akbar, M., Kaushik, A., Jamshed, M. A., Jung, H., & Hassan, S. A. J. I. C. M. (2025). LLM-based retrieval-augmented generation: a novel framework for resource optimization in 6G and beyond wireless networks. *63*(10), 60-67.

Zhou, H., & Chan, H. Y. J. a. p. a. (2026). ORCH: many analyses, one merge-a deterministic multi-agent orchestrator for discrete-choice reasoning with EMA-guided routing.

Zhu, J., Yan, L., Shi, H., Yin, D., & Sha, L. (2024). *ATM: Adversarial Tuning Multi-agent System Makes a Robust Retrieval-Augmented Generator*. https://doi.org/10.18653/v1/2024.emnlp-main.610